\documentclass{article} 
\usepackage{iclr2015,times}
\usepackage{hyperref}
\usepackage{url}

\usepackage{epsfig}
\usepackage{graphicx}
\usepackage{amsmath}
\usepackage{amssymb}

\usepackage{amsfonts}
\usepackage{amsthm}
\usepackage{subfig}

\newcommand{\eqn}[1]{Eqn.~\ref{eqn:#1}}
\newcommand{\fig}[1]{Fig.~\ref{fig:#1}}
\newcommand{\tab}[1]{Table~\ref{tab:#1}}
\newcommand{\secc}[1]{Section~\ref{sec:#1}}

\iclrfinalcopy 

\begin{document}
\headsep=20pt
\title{Training Convolutional Networks\\with Noisy Labels}

\author{
Sainbayar Sukhbaatar\\ 
Department of Computer Science, New York University\\ 
\texttt{sainbayar@cs.nyu.edu}
\And
Joan Bruna, Manohar Paluri, Lubomir Bourdev \& Rob Fergus \\
Facebook AI Research \\
\texttt{\{joanbruna,mano,lubomir,robfergus\}@fb.com}
}

\maketitle

\begin{abstract}
  The availability of large labeled datasets has allowed
  Convolutional Network models to achieve
  impressive recognition results. However, in many
  settings manual annotation of the data is impractical; instead our data has {\em noisy} labels, i.e.~there is some
  freely available label for each image which may or may not be
  accurate. In this paper, we explore the performance of
  discriminatively-trained Convnets when trained on such noisy data.
  We introduce an extra noise layer into the network which adapts the
  network outputs to match the noisy label distribution. The
  parameters of this noise layer can be estimated as part of the
  training process and involve simple modifications to current
  training infrastructures for deep networks. We demonstrate the
  approaches on several datasets, including large scale experiments on
  the ImageNet classification benchmark.
\end{abstract}

\vspace{-3mm}
\section{Introduction}
\vspace{-3mm}

In recent years, Convolutional Networks (Convnets) \citep{Lecun89,Lecun98} have shown impressive results on image classification tasks~\citep{krizhevsky2012imagenet,simonyan2014very}. However, this achievement relies on the availability of large amounts of labeled images, e.g.~ImageNet \citep{5206848}. Labeling images by hand is a laborious task and impractical for many problems. An alternative approach is to use labels that can be obtained easily, such as user tags from social network sites, or keywords from image search engines. The catch is that these labels are not reliable so they may contain misleading information that will subvert the model during training. But given the abundance of tasks where noisy labels are available, it is important to understand the consequences of training a Convnet on them, and this is one of the contributions of our paper.

\begin{figure}[h!]
\begin{center}
\includegraphics[width=13.5cm]{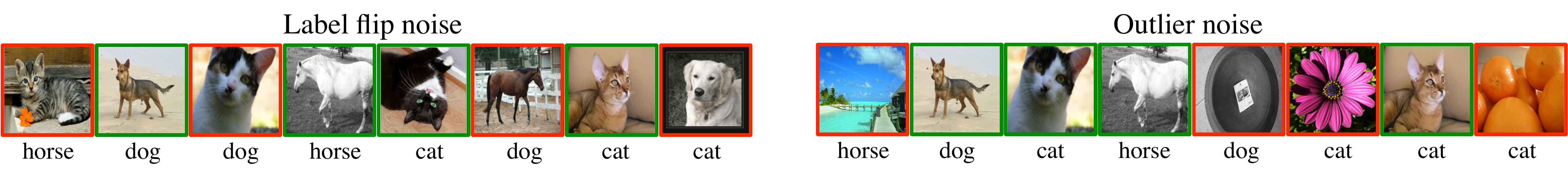}
\caption{A toy classification example with 3 classes, illustrating the two types of label noise encountered on real datasets. In the label flip case, the images all belong to the 3 classes, but sometimes the labels are confused between them. In the outlier case, some images are unrelated to the classification task but possess one of the 3 labels.}
\label{fig:teaser}
\end{center}
\vspace{-4mm}
\end{figure}

For image classification in real-world settings, two types of label noise dominate: (i) {\bf label flips}, where an example has erroneously been given the label of another class within the dataset and (ii) {\bf outliers}, where the image does not belong to any of the classes under consideration, but mistakenly has one of their labels. \fig{teaser} shows examples of these two cases.  We consider both scenarios and explore them on a variety of noise levels and datasets. Contrary to expectations, a standard Convnet model (from \citet{krizhevsky2012imagenet}) proves to be surprisingly robust to both types of noise. But inevitably, at high noise levels significant performance degradation occurs. 

Consequently, we propose a novel modification to a Convnet that enables it to be effectively trained on data with high level of label noise. The modification is simply done by adding a constrained linear ``noise'' layer on top of the softmax layer which adapts the softmax output to match the noise distribution. We demonstrate that this model can handle both label flip and outlier noise.  As it is a linear layer, both it and the rest of the model can be trained end-to-end with conventional back-propagation, thus automatically learning the noise distribution without supervision. The model is also easy to implement with existing Convnet libraries \citep{Cudaconv,Caffe,collobert2011torch7} and can readily scale to ImageNet-sized problems.  

\vspace{-5mm}
\section{Related Work}
\vspace{-3mm}
 
In any classification model, a degradation in performance is inevitable when there is noise in the training data \citep{nettleton2010study,1647654}. Especially, noise in labels is more harmful than noise in input features~\citep{zhu2004}. Label noise itself is a complex phenomenon. There are several types of noise on labels~\citep{6685834}. Also, noise source can be very different. For example, label noise can be caused by unreliable labeling by cheap and fast framework such as Amazon Mechanical Turk (http://www.mturk.com)~\citep{Ipeirotis:2010:QMA:1837885.1837906}, or noise can be introduced to labels intentionally to protect people privacy~\citep{INSR:INSR269}.

A simple approach to handle noisy labels is a data preprocessing stage, where labels suspected to be incorrect are removed or corrected \citep{barandela2000decontamination,brodley2011identifying}. However, a weakness of this approach is the difficulty of distinguishing informative hard samples from harmful mislabeled ones~\citep{Guyon:1996:DIP:257938.257955}. Instead, in this paper, we focus on models robust to presence of label noise.
The effects of label noise are well studied in common classifiers (e.g. SVMs, kNN, logistic regression), and robust variants have been proposed \citep{6685834,bootkrajang2012label}. Recently,~\citet{natarajan2013learning} proposed a generic unbiased estimator for binary classification with noisy labels. They employed a surrogate cost function that can be expressed by a weighted sum of the original cost functions, and gave theoretical bounds on the performance. 

Considering the recent success of deep learning~\citep{krizhevsky2012imagenet,taigman2014deepface,sermanet2013overfeat}, there is relatively little work on their application to noisy data.  In \citet{ICML2012Mnih_318} and \citet{675487}, noise modeling is incorporated to neural network in the same way as our proposed model. However, only binary classification is considered in \citet{ICML2012Mnih_318}, and \citet{675487} assumed symmetric label noise (i.e.~noise is independent of the true label). Therefore, there is only a single noise parameter, which can be tuned by cross-validation. In this paper, we consider multi-class classification and assume more realistic asymmetric label noise, which makes it impossible to use cross-validation to adjust noise parameters (for $k=10^3$ classes, there are $10^6$ parameters). 
Unsupervised pre-training of deep models has received much attention \citep{Hinton06,Erhan:2010:WUP:1756006.1756025}. Particularly relevant to our work is \citet{quoc} and \citet{Honglak09} who use auto-encoders to layer-wise pre-train the models. However, their performance has been eclipsed by purely discriminative Convnet models, trained on large labeled set~\citep{krizhevsky2012imagenet,simonyan2014very}.

In the paradigm we consider, all the data has noisy labels, with an unknown fraction being trustworthy. We do not assume the availability of any clean labels, e.g.~those provided by a human. This contrasts with semi-supervised learning (SSL) \citep{Zhu08}, where some fraction of the data has high quality labels but the rest are either unlabeled or have unreliable labels. Although closely related, in fact the two approaches are complementary to one another. Given a large set of data with noisy labels, SSL requires us to annotate a subset. But which ones should we choose? In the absence of external information, we are forced to pick at random. However, this is an inefficient use of labeler resources since only a fraction of points lie near decision boundaries and a random sample is unlikely to contain many of them. Even in settings where there is a non-uniform prior on the labels, picking informative examples to label is challenging. For example, taking high ranked images returned by an image search engine might seem a good strategy but is likely to result in prototypical images, rather than borderline cases. In light of this, our approach can be regarded as a natural precursor to SSL. By first applying our method to the noisy data, we can train a Convnet that will identify the subset of difficult examples that should be presented to the human annotator.

Moreover, in practical settings SSL  has several drawbacks that make it impractical to apply, unlike our method. Many popular approaches are based on spectral methods \citep{Zhu03,Zhou04,Zhu05} that have $O(n^3)$ complexity, problematic for datasets in the $n=10^6$ range that we consider. \citet{Fergus_nips09} use an efficient $O(n)$  spectral approach but make strong independence assumptions that may be unrealistic in practice. Nystrom methods \citep{Talwalkar08} can scale to large $n$, but do so by drastically sub-sampling first, resulting in a loss of fine structure within the problem. By contrast, our approach is $O(k^2)$ complexity, in the number of classes $k$, since we model the aggregate noise statistics between classes rather than estimating per-example weights. 

\vspace{-2mm}
\section{Label Noise Modeling}
\vspace{-1mm}
\subsection{Label Flip Noise}
\vspace{-1mm}
Let us first describe the scenario of \emph{label flip}.
Given training data $(\mathbf{x}_n, y^*_n)$ where $y^*$ denotes the true 
labels $\in {1,\dots K}$, we define a noisy label distribution $\tilde{y}$ given by 
$p(\tilde{y}=j | y^*=i) = q_{j,i}^*$,
parametrized by a $K \times K$ probability transition matrix $Q^*=(q^*_{ji})$.
We thus assume here that label flips are independent of $\mathbf{x}$. However, this model  
has the capacity to model asymmetric label noise distributions, as opposed to uniform 
label flips models \citep{675487}. For example, we can
model that a cat image is more likely to mislabeled as ``dog'' than ``tree''. 
The probability that an input $\mathbf{x}$ is labeled as $j$ in the noisy data can be computed using $Q^*$
\begin{align}
p(\tilde{y}=j | \mathbf{x}) &= \sum_i p(\tilde{y} = j | y^* = i)  p(y^* = i | \mathbf{x}) 
= \sum_i q^*_{ji}  p(y^* = i | \mathbf{x}) .
\label{eqA}
\end{align}

\begin{figure}[b!]
\vspace{-4mm}
\begin{center}
\subfloat[]{
\includegraphics[width=6cm]{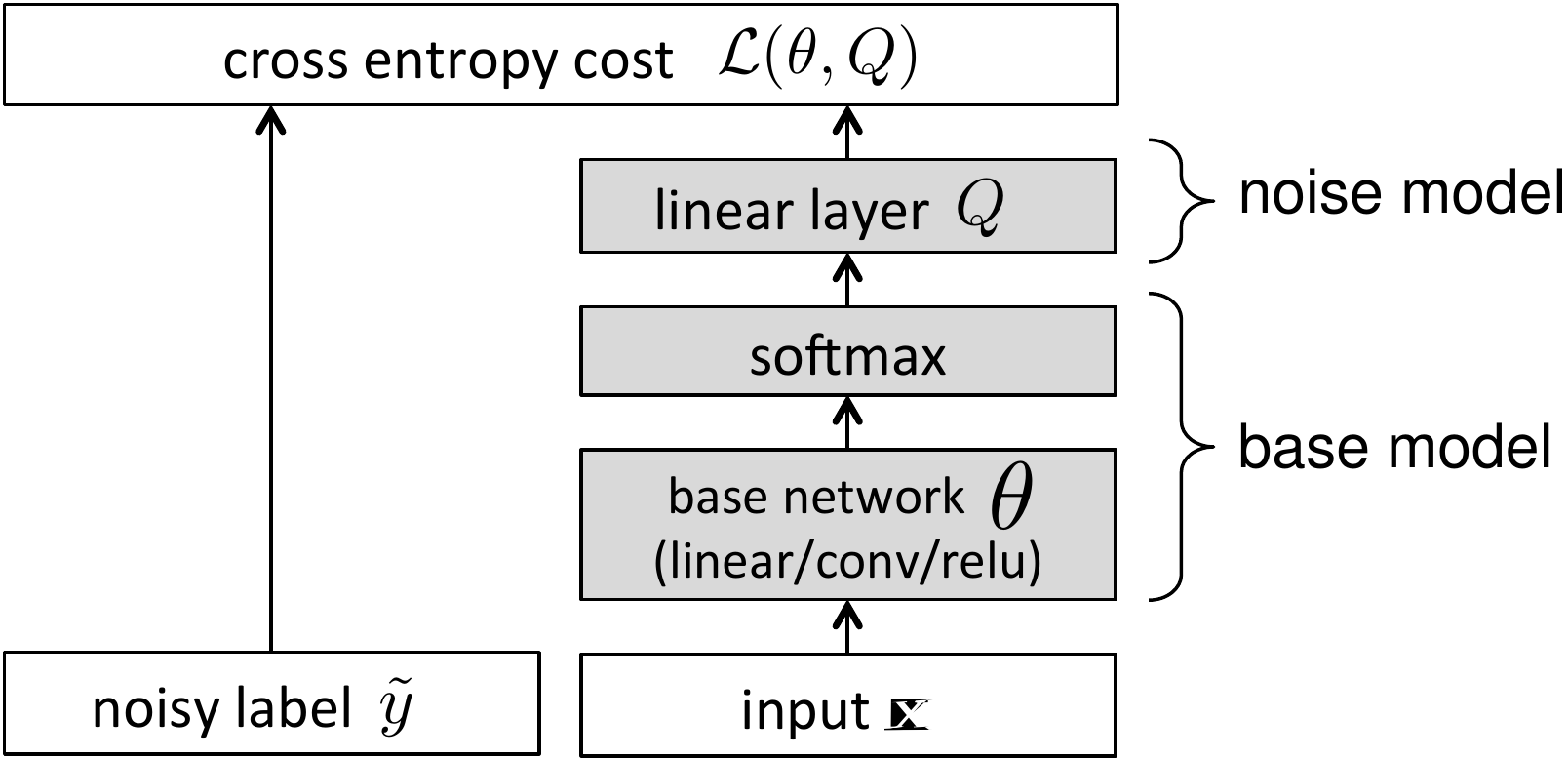}
\label{fig:model-bu}
}
\subfloat[]{
\includegraphics[width=7cm]{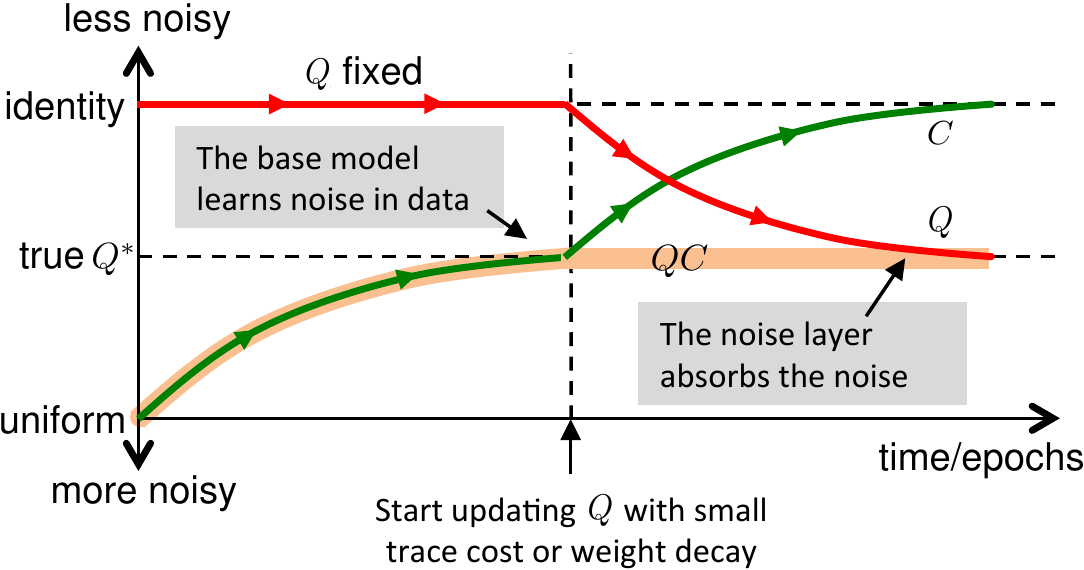}
\label{fig:training}
}
\caption{(a) Label noise is modeled by a constrained linear layer inserted between softmax and cost layers. Noise distribution $Q$ becomes the weight matrix of this layer. It changes output probabilities from the base model into a distribution that better matches the noisy labels. (b) The training sequence when learning from noisy data. The noise matrix ${Q}$ (red) is initially set to the identity, while the base model (green) is trained, inadvertently learning the noise in the data. Then we start updating ${Q}$ also (with regularization) and this captures the noise properties of the data, leaving the model to make ``clean'' predictions.}

\end{center}
\end{figure}

\vspace{-3mm}
In the same way, we can modify a classification model using a
probability matrix $Q$ that modifies its prediction to match the label
distribution of the noisy data. Let $\hat{p}(y^* | \mathbf{x},
\theta)$ be the prediction probability of true labels by the
classification model. Then, the prediction of the combined model will
be given by 
\vspace{-2mm}
\begin{align}
\hat{p}(\tilde{y}=j | \mathbf{x}, \theta,Q) = \sum_i q_{ji}  \hat{p}(y^* = i | \mathbf{x}, \theta) .
\label{eqA}
\end{align}
This combined model consist of two parts: the \emph{base model} parameterized by $\theta$ and the \emph{noise model} parameterized by $Q$. The combined model is trained by maximizing the cross entropy between the noisy labels $\tilde{y} $ and 
the model prediction given by Eqn.~\ref{eqA}. The cost function to minimize is 
\begin{align}
\mathcal{L}(\theta, Q) &= -\frac{1}{N}\sum_{n=1}^N \log{\hat{p}(\tilde{y}=\tilde{y}_{n} | \mathbf{x}_{n}, \theta,Q)} = -\frac{1}{N}\sum_{n=1}^N \log\left( \sum_i q_{\tilde{y}_{n}i} \hat{p}(y^*=i | \mathbf{x}_n, \theta)\right),
\label{bu-cost}
\end{align}
where $N$ is the number of training samples. 
However, the ultimate goal is to predict true labels $y^*$, not the noisy labels $\tilde{y}$.
This can be achieved if we can make the base model predict the true labels accurately. 
One way to quantify this is to use its confusion matrix $C=\{c_{ij}\}$ defined by
\begin{equation}
c_{ij} := \frac{1}{|S_j|} \sum_{n\in S_j} \hat{p}(y^*=i | \mathbf{x}_n, \theta),
\end{equation}
where $S_j$ is the set of training samples that have true label $y^*=j$. 
If we manage to make $C$ equal to identity, that means the base model perfectly predicts the true labels in training data. Note that this is the prediction before being modified by the noise model.
We can also define the confusion matrix $\tilde{C}=\{\tilde{c}_{ij}\}$ for the combined model in the same way
\begin{equation}
\tilde{c}_{ij} := \frac{1}{|S_j|} \sum_{n\in S_j} \hat{p}(\tilde{y}=i | \mathbf{x}_n, \theta, Q) .
\end{equation}
Using Eqn.~\ref{eqA}, it follows that $\tilde{C} = Q C$. Note that we cannot actually measure $C$ and $\tilde{C}$ in reality, unless we know the true labels.
Let us show that minimizing the training objective in Eqn.~\ref{bu-cost} forces the predicted distribution from the combined model to be as close as possible to the noisy label distribution of the training data, asymptotically. 
As $N \to \infty$, the objective in Eqn.~\ref{bu-cost}  becomes
\begin{align}
\label{eqj}
\mathcal{L}(\theta, Q) &= -\frac{1}{N}\sum_{n=1}^N \log{\hat{p}(\tilde{y}=\tilde{y}_{n} | \mathbf{x}_{n})}
= -\frac{1}{N} \sum_{k=1}^K \sum_{n \in S_k}  \log{\hat{p}(\tilde{y}=\tilde{y}_{n} | \mathbf{x}_{n},  y_n^* = k)} \nonumber \\ \hspace{-4mm}& 
\xrightarrow{N \to \infty} 
- \sum_{k=1}^K \sum_{i=1}^K  q_{ik}^* \log{\hat{p}(\tilde{y}=i |  \mathbf{x},  y^* = k)} 
\geq - \sum_{k=1}^K \sum_{i=1}^K q^*_{ik} \log q^*_{ik} ~, 
\end{align}
since $- \sum_k q_k^* \log \hat p_k \geq - \sum_k q_k^* \log q_k^* = H(q^*)$, and 
with equality in the last equation only when $\hat{p}(\tilde{y}=i | \mathbf{x},  y^* = k) = q^*_{ik}$.
In other words, the model tries to match the confusion matrix $\tilde{C}$ of the combined model to the true noise distribution $Q^*$ of the noisy data
\begin{equation}
\tilde{c}_{ik} =  1/|S_k| \sum_{n\in S_k} \hat{p}(\tilde{y}=i | \mathbf{x}_n, y_n^*=k) \to q^*_{ik} \implies \tilde{C} = Q C \to Q^*.
\label{qcq}
\end{equation}

If we know the true noise distribution $Q^*$ and it is non-singular, then from Eqn.~\ref{qcq}, setting $Q = Q^*$ would force $C$ to converge to identity. Therefore, training to predict the noisy labels using the combined model parameterized by $Q^*$ directly forces the base model to predict the true labels.
If the base model is a Convnet network with a softmax output layer,
then the noise model is a linear layer, constrained to be a
probability matrix, that sits on top of the
softmax layer, as shown in Figure~\ref{fig:model-bu}. The role of this
\emph{noise layer} is to implement Eqn.~\ref{eqA} on the output of the
softmax layer.  Therefore, the noise layer is a linear layer with no bias
and weights set to matrix $Q^*$.  Since this is the only modification
to the network, we can still perform back-propagation for training.

\vspace{-2mm}
\subsection{Learning the Noise Distribution}
\vspace{-1mm}
\label{sec:learnQ}
In the previous section, we showed that setting $Q=Q^*$ in the noise model is optimal for making the base model accurately predict true labels.
In practice, the true noise distribution $Q^*$ is often unknown to us. In this case, we have to infer it from the noisy data itself. 
Fortunately, the noise model is a constrained linear layer in our
network, which means its weights $Q$ can be updated along with other
weights in the network. This is done by back-propagating the
cross-entropy loss through the $Q$ matrix, down into the base
model. After taking a gradient step with the $Q$ and the model
weights, we project $Q$ back to the subspace of probability
matrices because it represents conditional probabilities.

Unfortunately, simply minimizing the loss in Eqn.~\ref{bu-cost} will not give us the desired solution.
From (\ref{eqj}), it follows that $Q C = \tilde{C} \to Q^*$ as the training progresses, where $\tilde{C}$ is the confusion matrix of the combined model and $Q^*$ is the true noise distribution of data. However, this alone cannot guarantee $Q \to Q^*$ and $C \to I_K$. For example, given enough capacity, the base model can start learning the noise
distribution and hence $C \to Q^*$, which implies that $Q \to I_K$. Actually, there are infinitely many solutions for $Q C = Q^*$ where $Q\neq Q^*$.

In order to force $Q \to Q^*$, we add a regularizer on the
probability matrix $Q$ which forces it to diffuse, such as a trace norm or
a ridge regression. This regularizer effectively transfers the label
noise distribution from the base model to the noise model,
encouraging $Q$ to converge to $Q^*$. Such
regularization is reminiscent of blind deconvolution algorithms.
Indeed, the noise distribution acts on our system by diffusing the
predictions of the base model. When the diffusion kernel is unknown,
it is necessary to regularize the ill-posed inverse problem by pushing
the estimates away from the trivial identity kernel \citep{Levin09}.

Under some strong assumptions, we can actually prove that $tr(Q)$ takes the smallest value only when $Q=Q^*$. Let us assume ${Q} C = Q^*$ holds true, and $Q$ and $Q^*$ have large diagonal elements (i.e. ${q}_{ii} > {q}_{ij}$ and $q_{ii}^* > q_{ij}^*$ for  $\forall i,j\neq i$). Then, it follows 
\begin{align}
\nonumber
 tr(Q^*) &= tr(Q C) = \sum_i (\sum_j q_{ij} c_{ji}) 
\le \sum_i (\sum_j q_{ii} c_{ji}) =\sum_i q_{ii}  (\sum_j c_{ji}) 
= \sum_i q_{ii} = tr(Q)~.
\end{align}
This shows that $tr(Q^*)$ is a lower bound for $tr(Q)$, and the equality will hold true only when $C = I_K$ and $Q=Q^*$. 
Therefore, minimizing $tr(Q)$ is a sensible way to make the base model accurately predict clean labels. Although the above proof is true under strong assumptions, we  show empirically that it works well in practice. Also, we use weight decay on ${Q}$ instead of minimizing $tr(Q)$ in practice since it is already implemented in most deep learning packages and has similar effect of diffusing $Q$. 

Let us finally describe the learning procedure, illustrated in Figure
\ref{fig:training}.  In the beginning of training, $QC$ and $Q^*$ are
very different in general, and the confusion matrix $C$ does not
necessarily have large elements on its diagonal. This makes learning
$Q$ difficult. Therefore, we fix $Q=I_K$ at the start of training. This is illustrated in the left part of Figure \ref{fig:training}.  At that point, the base model
could have learned the noise in the training data (i.e. $C=Q^*$),
which is not what we want.  Therefore, we start updating $Q$ along
with the rest of the network, using weight decay to push $Q$ away from
the identity and towards $Q^*$.  As ${Q}$ starts to diffuse, it starts
absorbing the noise from the base model, thus making the base model
more accurate.  However, too large weight decay would make ${Q}$ more
diffused than the true $Q^*$, which may hurt the performance. In case there is no clean data, we cannot tune this weight decay parameters with validation. In the experiments in this paper, we fix the weight decay parameter for $Q$ to $0.1$ (in some cases where the training classification cost significantly increased because of the weight decay on $Q$, we used smaller weight decay parameter). If we want to make prediction or test the model on clear data, the noise layer should be
removed (or set to identity $I$).

\vspace{-2mm}
\subsection{Outlier Noise}
\vspace{-2mm}
\label{sec:outlier}
Another important setting is the case where some training samples do not belong to any of the existing signal classes.
In that case, we can create an additional ``outlier'' class, which enables us to apply the previously described noise model.

Let $K$ be the number of the existing classes. Then, the base network should output $K+1$ probabilities now, where the last one represents the probability of a sample being an 
outlier. If the labels given to outlier samples are uniformly
distributed across classes, then the corresponding noise distribution $Q^*$ becomes a $K+1\times K+1$ matrix
\begin{equation}
Q^* = 
{\tiny
\begin{pmatrix}
1 & 0 & \cdots & 0 & 1/K \\
0 & 1 & \cdots & 0 & 1/K \\
\vdots & \vdots & \ddots & \vdots & \vdots \\
0 & 0 & \cdots & 1 & 1/K \\
0 & 0 & \cdots & 0 & 0 \\
\end{pmatrix}
}.
\end{equation}
Unfortunately, this matrix is singular and would map two different network outputs $y_1=(0,...,0,1)^T$ and $y_2=(1/K,...,1/K,0)$ to the exact same point. 
A simple solution to this problem is to add some extra outlier images with label $K+1$ in the training data, 
which would make $Q^*$ non-singular (in most cases, it is cheap to obtain such extra outlier samples). Now, the noise distribution becomes
\vspace{-2mm}
\begin{equation}
Q^* = 
{\tiny
\begin{pmatrix}
1 & 0 & \cdots & 0 & (1-\alpha)/K \\
0 & 1 & \cdots & 0 & (1-\alpha)/K \\
\vdots & \vdots & \ddots & \vdots & \vdots \\
0 & 0 & \cdots & 1 & (1-\alpha)/K \\
0 & 0 & \cdots & 0 & \alpha \\
\end{pmatrix}
} ~, \hspace{4mm} \text{where} \:\: \alpha= \frac{|\text{outliers labeled ``} K+1\text{''}|}{|\text{total outliers}|}.
\label{eqn:outlierQ}
\end{equation}
Note that in this setting there is no learning: $Q$ matrix is fixed
to $Q^*$ given in \eqn{outlierQ}. The fraction of outliers in the
training set, required to compute $\alpha$, is a hyper-parameter
that must be set manually (since there is no principled way to
estimate it). However, we experimentally demonstrate that the algorithm
is not sensitive to the exact value. 

\vspace{-3mm}
\section{Experiments}
\vspace{-2mm}
In this section, we empirically examine the robustness of deep
networks with and without noise modeling. First, we perform controlled
experiments by deliberately adding two types of label noise to clean
datasets: label flip noise and outlier noise. Then we show more realistic experiments using two datasets with inherent label noise, where we do not know the true distribution of noisy labels. 
\vspace{-2mm}
\subsection{Data and Model Architecture}
\vspace{-2mm}
We use three different image classification datasets in our
experiments. The first dataset is the Google street-view house number dataset
(SVHN)~\citep{netzer2011reading}, which consists of 32x32 images of
house number digits captured from Google Streetview. It has about 600k
images for training and 26k images for testing. The second one is more
challenging dataset CIFAR10~\citep{krizhevsky2009learning}, a subset of
80 million Tiny Images dataset~\citep{4531741} of 32x32 natural images
labeled into 10 object categories. There are 50k training images and
10k test images. The last dataset is ImageNet~\citep{5206848}, a large
scale dataset of 1.2M images, labeled with 1000 classes.
 For all datasets, the only preprocessing step is mean subtraction,
except for SVHN where we also perform contrast normalization. For
ImageNet, we use data augmentation by taking random crop of 227x227 at
random locations, as well as horizontal flips with probability 0.5.

For SVHN and CIFAR-10 we use the model architecture and
hyperparameter settings given by the CudaConv~\citep{Cudaconv} configuration file
\texttt{layers-18pct.cfg}, which implements a network with three
convolutional layers. These settings were kept fixed for all
experiments using these datasets. For ImageNet, we use the model architecture
described in \citet{krizhevsky2012imagenet} (AlexNet). 

\vspace{-2mm}

\subsection{Label Flip Noise}
\vspace{-1mm}
 
\begin{figure}[tb]
\begin{center}
\subfloat[]{
\includegraphics[width=4.0cm,trim = 5mm 0mm 0mm 0mm]{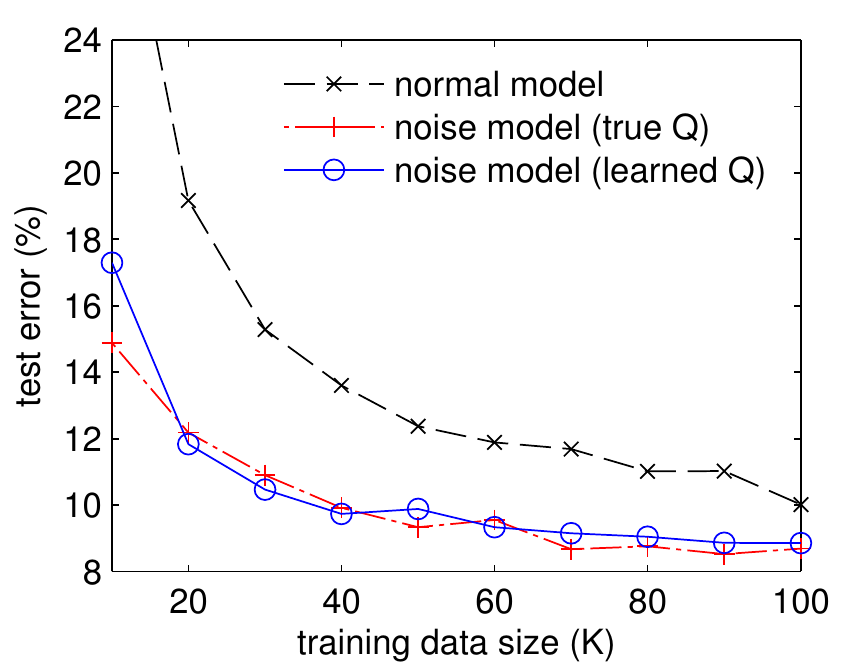} 
\label{fig:svhn-err}
}
\subfloat[]{
\includegraphics[width=4.0cm,trim = 5mm 0mm 0mm 0mm]{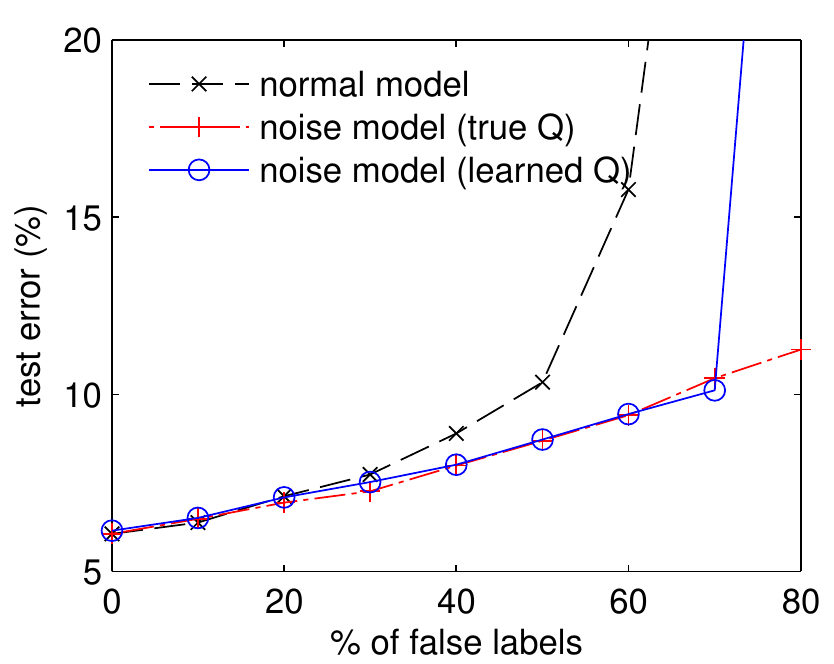}
\label{fig:svhn-err2}
}\quad
\subfloat[]{
\includegraphics[width=5.0cm]{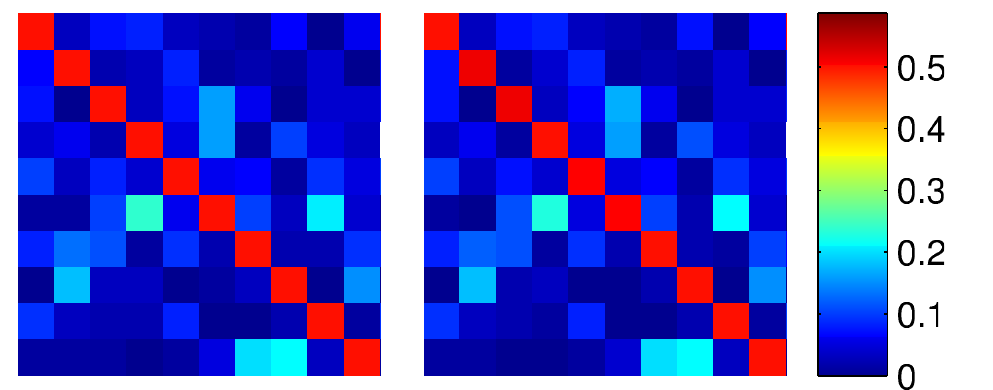}
\label{fig:svhn-err3}
}
\vspace{-2mm}
\caption{(a) Test errors on SVHN dataset when the noise level is 50\%
  for differing overall training set sizes. (b)
  Test errors when trained on 100k samples, as the noise level
  varies. Note that the performance for learned $Q$ is very close to a
  model trained with $Q$ fixed to the true noise distribution $Q^*$.
  (c) The ground truth noise distribution $Q^*$ (left) and ${Q}$ learned from noisy data (right).}
\end{center}
\vspace{-3mm}
\end{figure}

We synthesize noisy data from clean data by stochastically changing some
of the labels:  an original label $i$ is randomly changed to $j$ with
fixed probability $q_{ji}^*$. Figure~\ref{fig:svhn-err3}  shows the noise distribution
$Q^*=\{q_{ji}^*\}$ used in our experiments. We can alter this distribution
by changing the probability on the diagonal to generate datasets
with different overall noise levels. The labels of test images are left
unperturbed. \newline
\noindent {\bf SVHN:} 
When training a noise model with SVHN data, we fix $Q$ to the
identity for the first five epochs. Thereafter, ${Q}$ is updated
with weight decay 0.1 for 95
epochs. Figure~\ref{fig:svhn-err} and \ref{fig:svhn-err2} shows the test errors for
different training set sizes and different noise levels. These plots
show the normal model coping with up to 30$\%$ noise, but then
degrading badly. By contrast, the addition of our noise layer allows
the model to operate with up to 70$\%$. Beyond this, the method breaks
down as the false labels overwhelm the correct ones. Overall, the
addition of the noise model consistently achieves better accuracy,
compared with a normal deep network. The figure also shows error rates for
a noise model trained using the true noise distribution $Q^*$. We see
that it performs as well as the learned ${Q}$, showing that our
proposed method for learning the noise distribution from data is
effective. Figure~\ref{fig:svhn-err3} shows an example of a learned ${Q}$
alongside the ground truth $Q^*$ used to generate the noisy data. We
can see that the difference between them is negligible.

\begin{figure}[h]
\begin{center}
\scriptsize
\begin{tabular}{cccc}
\underline{Without noise model} & \underline{With learned noise model} &
\underline{Without noise model} & \underline{With learned noise model} \\ 
\includegraphics[scale=0.45,trim = 1mm 0mm 17mm 0mm,clip=true]{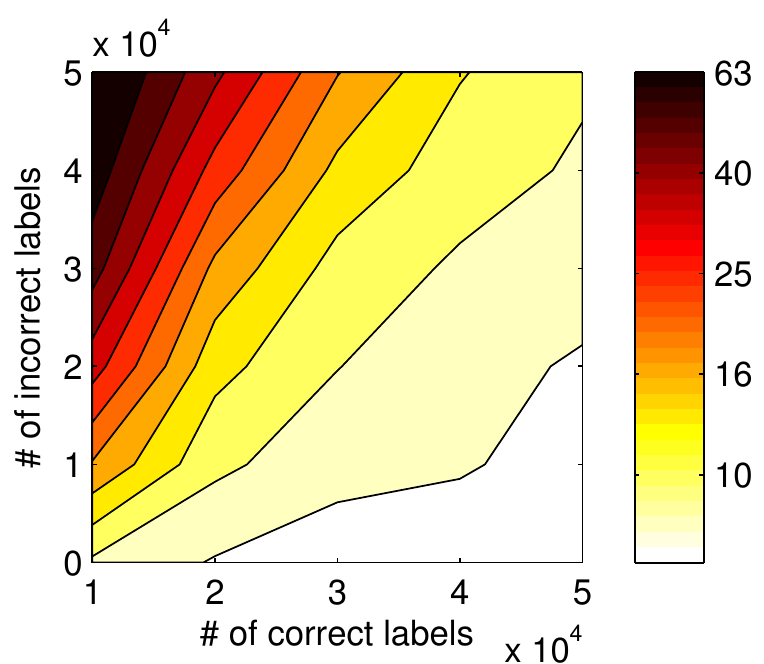} &
\includegraphics[scale=0.45,trim = 6mm 0mm 0mm 0mm,clip=true]{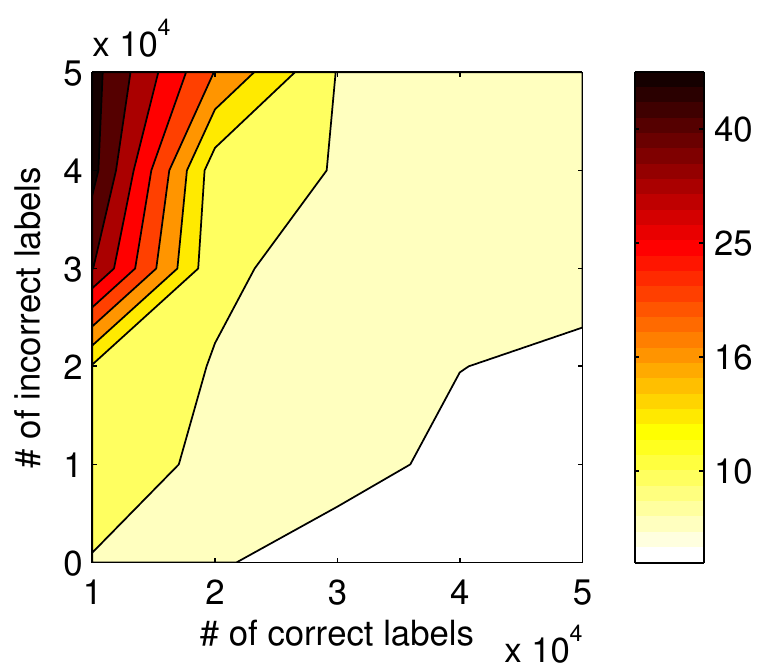}
&
\includegraphics[scale=0.45,trim = 1mm 0mm 17mm 0mm,clip=true]{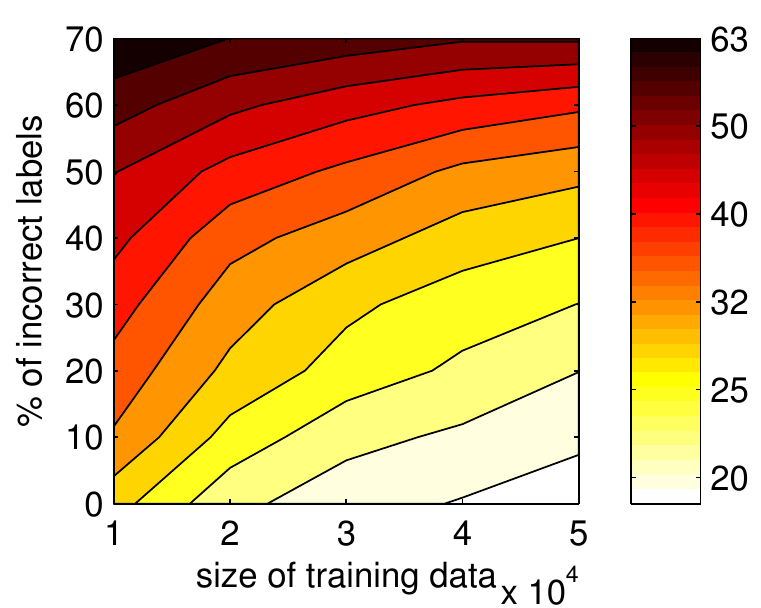} &
\includegraphics[scale=0.45,trim = 6mm 0mm 0mm 0mm,clip=true]{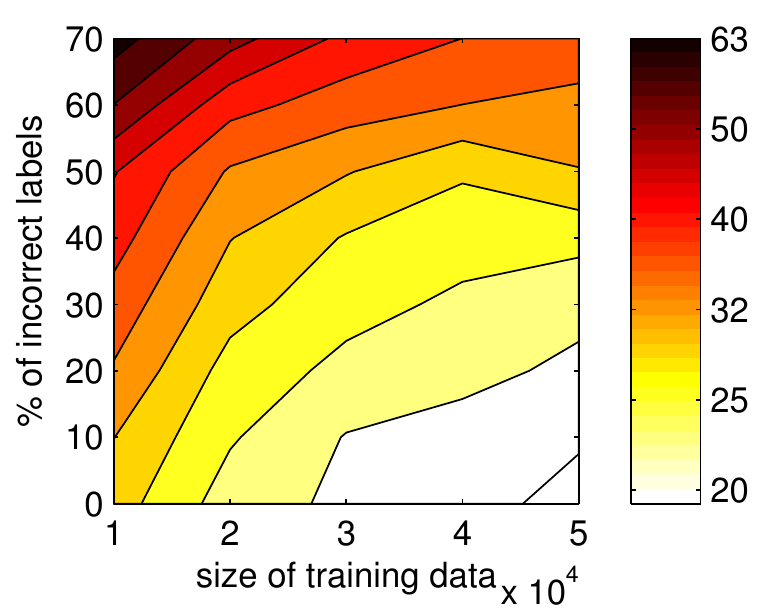} \\
\multicolumn{2}{ c }{(a) SVHN} & \multicolumn{2}{ c }{(b) CIFAR10}
\end{tabular}
\caption{The noise model is compared to the baseline model on different amount of training data and varying noise levels. The plots show test errors (\%), where brighter color indicates better accuracy.}
\label{fig:svhn-grid}
\end{center}
\vspace{-6mm}
\end{figure}

Figure~\ref{fig:svhn-grid}a shows the effects of label flip noise in
more detail. The color in the figure shows the test errors (brighter
means better), and the contour lines indicates the same accuracy.
Without the noise model, the performance drops quickly as the number
of incorrect labels increases. In contrast, the convnet with the noise
model shows greater robustness to incorrect labels.

\vspace{-2mm}
\noindent {\bf CIFAR-10:}
We perform the same experiments as for SVHN on the CIFAR-10 dataset,
varying the training data size and noise level. We fix ${Q}$ to
identity for the first 30 epochs of training and then run for another
70 epochs updating ${Q}$ with weight decay 0.1. The results are
shown in Figure~\ref{fig:svhn-grid}b. Again, using the noise model is
more robust to label noise, compared to the unmodified model. The
difference is especially large for high noise levels and large
training sets, which shows the scalability of the noise model.

\noindent {\bf ImageNet:}
In this experiment, we deliberately flip half of the training labels
in ImageNet dataset to test the scalability of our noise model to a
1000 class problem. We
explore two different noise distributions: (i) random and (ii)
adversarial, applied to the ImageNet 2012 training set. In the random case, labels are flipped with a non-uniform
probability using a pre-defined matrix $Q^*$, which has around 10
thousand non-zero values at random off-diagonal locations. I.e~for
each class, 50\% of the labels are correct, with the other 50\% being distributed over 10 other
randomly chosen classes. In the adversarial case, we use a noise
distribution where labels are changed to
other classes that are more likely to be confused with the true
class (e.g. simulating human mislabeling). First, we train a normal convnet on clean data and measure its
confusion matrix. This matrix gives us a good
metric of which classes more likely to be confused. Then this matrix
is used for constructing $Q^*$ so that 40\% of labels are randomly
flipped to other similar labels.

Using the AlexNet architecture \citep{krizhevsky2012imagenet}, we train three models for each noise
case: (i) a standard model with no noise layer; (ii) a model with a
learned $Q$ matrix and (iii) a model with $Q$ fixed to the ground
truth $Q^*$. 

Table~\ref{imgnet} shows top-1 classification error on the ImageNet
2012 validation set for models trained on the random noise
distribution. It is clear that the noise hurts performance
significantly. The model with learned $Q$ shows a clear gain (8.5\%) over the
unaltered model, but is still 3.8\% behind the model that used the
ground truth $Q^*$. The learned $Q$ model is superior to training an
unaltered model on the subset of clean labels, showing that the noisy
examples carry useful information that can be accessed by our model. 
Table~\ref{imgnet2} shows errors for the adversarial noise
situation. Here, the overall performance is worse (despite a lower
noise level than Table~\ref{imgnet}, but the learned noise model is
still superior to the unaltered model.

\begin{table}[htdp]
\begin{center}
\subfloat[random label flip noise]{
\begin{tabular}{|p{1.6cm}|p{1.2cm}|p{0.8cm}|p{1cm}|}
\hline
Noise model & Training size & Noise \%  & Valid. Error \\ \hline \hline
None & 1.2M & 0 & 39.8\% \\ \hline
None & 0.6M & 0 & 48.5\% \\ \hline
None & 1.2M & 50 & 53.7\% \\ \hline
Learned $Q$ & 1.2M & 50 & 45.2\% \\ \hline
True $Q^*$ & 1.2M &50 & 41.4\% \\ \hline
\end{tabular}
\label{imgnet}
}\quad
\subfloat[adversarial label flip noise]{
\begin{tabular}{|p{1.6cm}|p{1.2cm}|p{0.8cm}|p{1cm}|}
\hline
Noise model & Training size & Noise \% & Valid. error \\ \hline \hline
None & 1.2M & 40 & 50.5\% \\ \hline
Learned $Q$  & 1.2M & 40 & 46.7\% \\ \hline
True $Q^*$ & 1.2M & 40 & 43.3\% \\ \hline
\end{tabular}
\label{imgnet2}
}
\vspace{-3mm}
\caption{Effect of label flip noise using the ImageNet dataset}
\vspace{-6mm}
\end{center}
\end{table}

\vspace{-3mm}
\subsection{Outlier Noise}
\vspace{-2mm}
{\bf CIFAR-10:} Here, we simulate outlier noise by deliberately
polluting CIFAR10 training data with randomly chosen images from the
Tiny Images dataset~\citep{4531741}. Those random images can be
considered as outliers because Tiny Images dataset covers $\sim$75,000
classes, thus the chance of belonging to 10 CIFAR classes is very
small. Our training data consists of a random mix of inlier images with true labels
and outlier images with random labels (n.b.~the model has no knowledge of which
are which). As described in \secc{outlier}, the outlier model requires
a small set of known outlier images. In this case, we use 10k examples
randomly picked from Tiny Images. 
For testing, we use the original (clean) CIFAR10 test data.

\begin{figure}[b!]
\vspace{-3mm}
\begin{center}
\mbox{
\includegraphics[width=6.5cm]{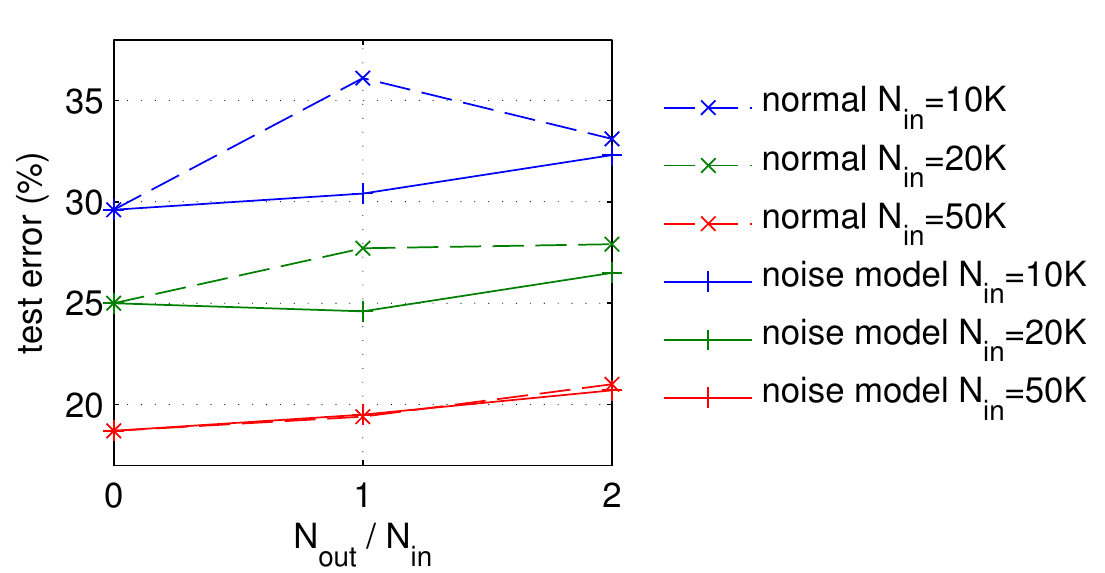}
\includegraphics[width=6.5cm]{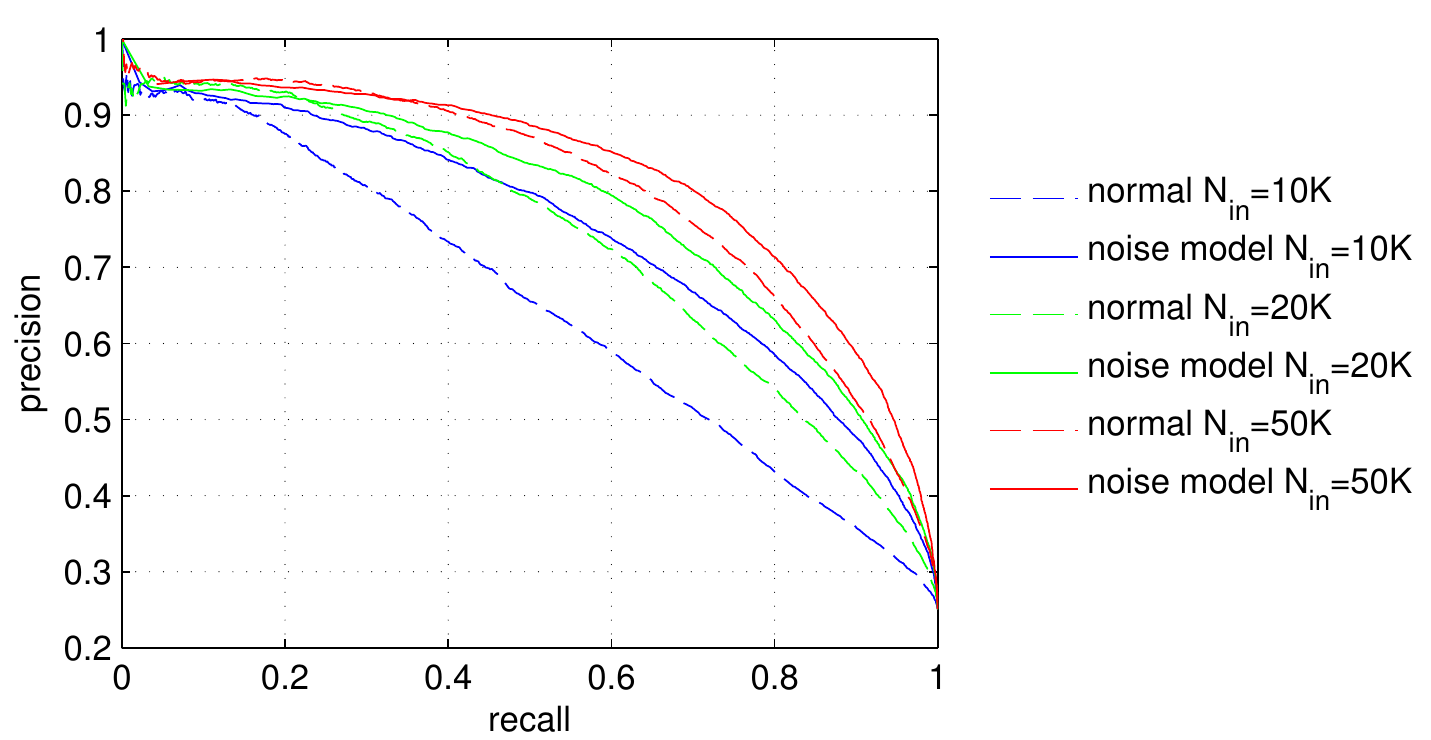}
}
\vspace{-2mm}
\caption{CIFAR-10 outlier experiments. Left: The effect of the outlier noise on the classification performance on CIFAR10 with and without the noise model. Here, $N_{in}$ and $N_{out}$ are the number of inlier and outlier images in the training data, respectively. Right: Precision recall curve for detecting inliers in test data.}
\label{cifar10backeffect}
\end{center}
\end{figure}

\begin{figure}[h!]
\begin{center}
\includegraphics[width=7.0cm]{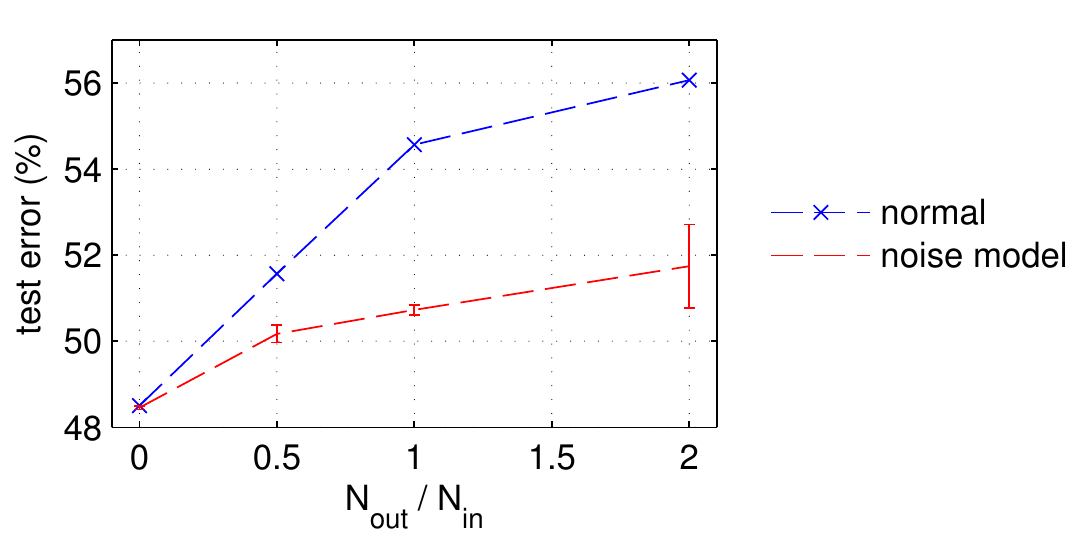}
\caption{ImageNet outlier experiments with varying ratios of outlier/inliers.}
\label{fig:INout}
\end{center}
\end{figure}

Figure~\ref{cifar10backeffect}(left) shows the classification performance of
a model trained on different amounts of inlier and outlier
images. Interestingly, a large amount of outlier noise does not
significantly reduce the accuracy of the normal model without any
noise modeling. Nevertheless, when the noise model is used the effect
of outlier noise is reduced, particularly for small training sets. In
this experiment, we set hyper-parameter $\alpha$ in
Eqn.~\ref{eqn:outlierQ} using the true number of outliers but, as
shown in \fig{INout}, model is not sensitive to the precise value.

Figure~\ref{cifar10backeffect}(right) explores the ability of the
trained models to distinguish inlier from outlier in a held-out noisy
test set. For the normal model, we use the entropy of the softmax
output as a proxy for outlier confidence. For our outlier model, we
use the confidence of the $k+1_{th}$ class output. The figure shows
precision recall curves for both models trained varying training set
sizes with 50\% outliers. The average precision for the outlier model is consistently
superior to the normal model.    

{\bf ImageNet:} We added outlier noise
to ImageNet dataset (1.2M images, 1K categories) using 
images randomly chosen from the entire ImageNet Fall 2011 release
(15M images, 22K categories). Each outlier images is randomly labeled as one of 1K
categories and added to the training set.

We fix the number of inlier images $N_{in}$ to be $0.6M$ (half of the
original ImageNet training data). We increase the number of outlier images
up to $N_{out}=1.2M$, mixed into the training data. For training the noise
model, we added 20K outlier images, labeled as
outlier to the training data. We trained a normal AlexNet model on the
data, along with a version using the outlier model. 
We used three different $\alpha$ values (see Eqn.~\ref{eqn:outlierQ})
in these experiments. One run used $\alpha$ set using the true percentage
of outliers in the training data. The other two runs perturbed this
value by $\pm 15\%$, to explore the sensitivity of the noise model to
this hyper-parameter. The
results are shown in \fig{INout} (the error bars show $\pm 1\sigma$)
for differing amounts of outliers. Although the normal deep
network is quite robust to large outlier noise, using the noise model
further reduces the effect of noise.

\subsection{Real Label Noise}

{\bf Tiny Images:}
The Tiny Images dataset~\citep{4531741} is a realistic source of
noisily labeled data, having been gathered from Internet search
engines. We apply our outlier model to a 200k subset,
taken from the 10 classes from which the CIFAR-10 dataset was
extracted. The data contains a mix of inlier images, as well as
totally unrelated outlier images. After a cursory examination of the
data, we estimated the outlier fraction to be $0.5$. Using this ratio,
along with 5k known outliers (randomly drawn Tiny Images), $\alpha$
was set to $0.05$. For evaluation we used the CIFAR-10 test
set\footnote{We carefully removed from the training set any images
  that were present in the CIFAR-10 test set.}. Training on this data
using an unmodified convnet produced a test error of 19.2\%. A second model, trained with
an outlier layer, gave a test error of 18.8\%, a relative gain of
2.1\%.

{\bf Web Images + ImageNet:} 
We also apply our approach to a more challenging noisy real world
problem based around the 1000 classes used in ImageNet. We collected a
new noisy {\em web image}  dataset, using each of the thousand
category names as input into Internet image search engines. All
available images for each class were downloaded, taking care to delete
any that also appeared in the ImageNet dataset. This dataset
averages 900 examples/class for a total of 0.9M images. The noise
is low for the highly ranked images, but significant for the later
examples of each class. The precise noise level is unknown, but after browsing some of the images we set
$\alpha=0.08$, assuming 20\% of images are outlier. We trained an AlexNet model \citep{krizhevsky2012imagenet} with and without the
noise adaption layer on this web image
dataset and evaluated the performance on the ImageNet 2012 validation set. 
One complication is that the distribution of inliers in the web
images differs somewhat from the ImageNet evaluation set, creating a
problem of domain shift. To reduce this effect (so that the noise adaptation
effects of our approach can be fairly tested), we added 0.3M ImageNet
training images to our web data. This ensures that the model learns
a representation for each class that is consistent with the test data.

\tab{imagenet} shows three Alexnet models applied to the data: (i) an
unaltered model, (ii) a model with learned label-flip noise matrix (weigh decay parameter for $Q$ set to 0.02 because larger value significantly increased the training classification cost) and
(iii) a model with an outlier noise matrix (with $\alpha$ set to
0.08). The results show the label-flip model boosting performance by 0.6\%.

\begin{table}[htdp]
\begin{center}
{
\begin{tabular}{|l|c|}
\hline
Method & Valid. error \\ \hline \hline
Normal Convnet & 48.8\% \\ \hline
Label-flip model & 48.2\% \\ \hline
Outlier model & 48.5\% \\ \hline
\end{tabular}}
\caption{Evaluation on our real-world Web image + ImageNet noisy dataset.}
\label{tab:imagenet}
\end{center}
\vspace{-4mm}
\end{table}

\vspace{-2mm}
\section{Conclusion}
\vspace{-1mm}
In this paper we explored how convolutional networks can be trained
on data with noisy labels. We proposed two simple models for improving noise
robustness, focusing different types of noise. We explored both approaches
in a variety of settings: small and large-scale datasets, as well as
synthesized and real label noise. In the former case, both
approaches gave significant performance gains over a standard
model. On real data, then gains were smaller. However, both approaches can
be implemented with minimal effort in existing deep learning
implementations, so add little overhead to any training procedure. 

\small
\bibliography{paper}
\bibliographystyle{iclr2015}

\end{document}